\documentclass[conference]{IEEEtran}
\IEEEoverridecommandlockouts
\usepackage{cite}
\usepackage{amsmath,amssymb,amsfonts}
\usepackage{algorithmic}
\usepackage{graphicx}
\usepackage{textcomp}
\usepackage{xcolor}

\usepackage{booktabs}
\def\BibTeX{{\rm B\kern-.05em{\sc i\kern-.025em b}\kern-.08em
    T\kern-.1667em\lower.7ex\hbox{E}\kern-.125emX}}
\begin{document}

\title{DiffMagicFace: Identity Consistent Facial Editing of Real Videos}


\author{
\IEEEauthorblockN{
Huanghao Yin\IEEEauthorrefmark{1},
Shenkun Xu\IEEEauthorrefmark{2},
Kanle Shi\IEEEauthorrefmark{2},
Junhai Yong\IEEEauthorrefmark{1},
Bin Wang\IEEEauthorrefmark{1}
}
\IEEEauthorblockA{\IEEEauthorrefmark{1}Tsinghua University}
\IEEEauthorblockA{\IEEEauthorrefmark{2}Kuaishou Technology}
\IEEEauthorblockA{\IEEEauthorrefmark{1}yinhh22@mails.tsinghua.edu.cn, yongjh@tsinghua.edu.cn, wangbins@tsinghua.edu.cn}
\IEEEauthorblockA{\IEEEauthorrefmark{2}xushenkun@kuaishou.com, shikanle@kuaishou.com}
}

\maketitle

\begin{figure*}[!t]
\centering
\includegraphics[width=\linewidth]{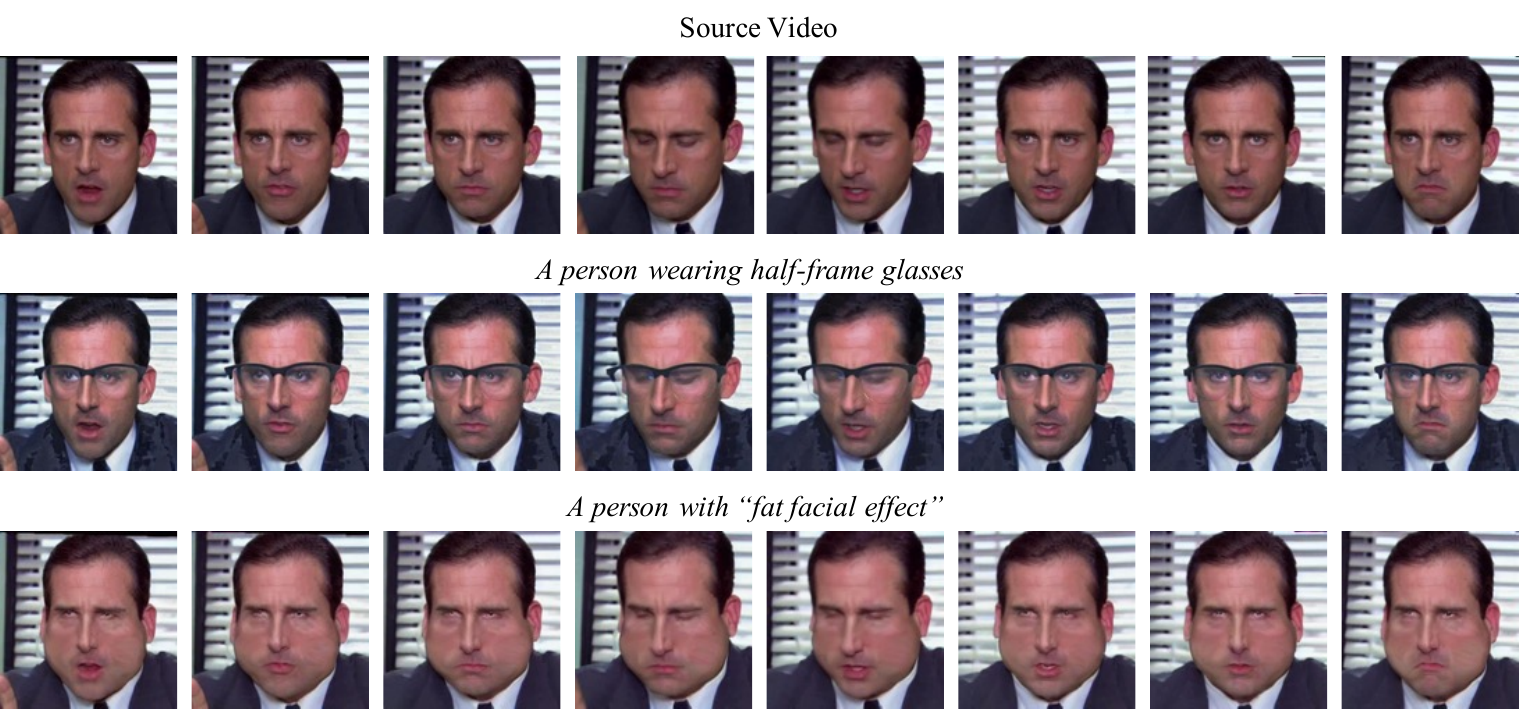}
\caption{\textbf{Identity consistent face video editing. }Our method achieves face video editing on a specific subject (e.g., half-frame glasses, or fat facial effect) using a text prompt, while preserving the identity of the source video and maintaining high consistency among frames.}
\label{t1}
\end{figure*}

\begin{abstract}
Text-conditioned image editing has greatly benefitted from the advancements in Image Diffusion Models. However, extending these techniques to facial video editing introduces challenges in preserving facial identity throughout the source video and ensuring consistency of the edited subject across frames. In this paper, we introduce DiffMagicFace, a unique video editing framework that integrates two fine-tuned models for text and image control. These models operate concurrently during inference to produce video frames that maintain identity features while seamlessly aligning with the editing semantics. To ensure the consistency of the edited videos, we develop a dataset comprising images showcasing various facial perspectives for each edited subject. The creation of a data set is achieved through rendering techniques and the subsequent application of optimization algorithms. Remarkably, our approach does not depend on video datasets but still delivers high-quality results in both consistency and content. The excellent effect holds even for complex tasks like talking head videos and distinguishing closely related categories. The videos edited using our framework exhibit parity with videos that are made using traditional rendering software. Through comparative analysis with current state-of-the-art methods, our framework demonstrates superior performance in both visual appeal and quantitative metrics.
\end{abstract}


\section{Introduction}
The field of facial video editing has seen remarkable advancements due to the advent of large-scale text-to-image generative models, such as DALL-E~\cite{ramesh2022dalle2}, Imagen~\cite{saharia2022imagen}, and Stable Diffusion~\cite{rombach2022high}. These models have shifted the paradigm from earlier Generative Adversarial Network (GAN)-based approaches~\cite{tzaban2022stitch,abdal2022video2stylegan,skorokhodov2022styleganv, kim2021exploiting, cheng2020sequential}, offering a new level of flexibility in editing by leveraging textual prompts. Despite their promise, there remains a significant gap in the literature on their application to facial video editing, a niche where maintaining consistency of identity between frames is paramount.

Among the sparse contributions~\cite{zhang2023sine,brooks2023instructpix2pix,kim2023dva,gal2022textual,ceylan2023pix2video,molad2023dreamix,liu2023videop2p,wu2023tune}, the method by Kim ~\textit{et al.}~\cite{kim2023dva}, which minimizes directional CLIP loss for semantic editing, stands out. However, this approach often results in edits that are not significantly distinct from the original video, due to the limited emphasis on identity preservation during optimization and the semantic ambiguity of direct latent space editing.

Some other influential image-editing works, such as InstructPix2Pix~\cite{brooks2023instructpix2pix} and Textual Inversion~\cite{gal2022textual}, are capable of generating noticeable modifications to facial images using text prompts. However, these methods suffer from the inability to preserve the original human identity of the source facial image, and they are unable to maintain the same editing props when generating multiple frames.

Our work introduces a novel framework that addresses the above two shortcomings employing a dual-model strategy for text and image control of the facial video editing task. This framework ensures clear, distinct edits from textual descriptions while maintaining the human identity and the same editing subject throughout the video. By fine-tuning a specialized text control model alongside an image control model, we enable the production of complex visual effects directly from simple text prompts, a significant leap over existing methodologies.

In the domain of facial video editing applications, there has been a significant rise in popularity of short video special effects, commonly referred to as ``magic-face'' on various platforms. These ``magic-face'' aim to enhance the expression effect of the facial frame through the use of elements such as makeup, decorations, and animated graphics. There is a growing demand for ``magic-face'' effects to provide users with more sophisticated and intricate features. 

Nevertheless, existing pre-trained T2I diffusion models~\cite{ramesh2022dalle2,saharia2022imagen,rombach2022high} are wholly lack of these ``magic-face'' semantics since they are not within the data distribution of the training set. In order to obtain expressive results like ``magic-face'' in video editing, training a customized dataset is necessary.
Our work provides a method of generating datasets in the style of ``magic-face'' with a rendering software that can be applied in fine-tuning diffusion models. 

This paper introduces \textbf{\textit{DiffMagicFace}}, a model-based approach for generating ``magic-face'' effects in videos. This approach exhibits its ability to execute complex visual transformations while preserving the original human identity and ensuring editing subject's continuity across frames. This approach outperforms previous methods by offering broader categorical flexibility and maintaining higher fidelity in content and identity.

In summary, our contributions are threefold:
\begin{itemize}
\item We introduce a hybrid framework that combines diffusion models for text and image control, skillfully balancing semantic editing with identity preservation.
\item Our approach uniquely creates a paired and edited training dataset without relying on specific video datasets, utilizing advanced rendering techniques and an image captioning model.
\item We set a new benchmark in facial video editing, consistently delivering state-of-the-art results in maintaining identity integrity across edits.
\end{itemize}

\section{Related Work}
\subsection{Video editing.}
When editing real videos, preserving temporal consistency is imperative. Initially, GAN-based video editing methods \cite{tzaban2022stitch,abdal2022video2stylegan,skorokhodov2022styleganv,yao2021latent} predominantly prioritize the manipulation of global video features, encompassing artistic style transfer, colorization, and image enhancement. To promote temporal consistency in these techniques, their training process depends on video data that incorporates a temporal loss metric, quantified as the warping error between the resulting frames. Subsequently, recent diffusion-based video generating and editing methods \cite{molad2023dreamix,ho2022imagenvideo,singer2022makeavideo,ceylan2023pix2video,liu2023videop2p} have exhibited notable success. A notable advantage of diffusion-based approaches lies in their ability to edit images based on a wide range of textual prompts, as opposed to relying on fixed categories present in datasets. However, it is important to note that many of these approaches do not place a primary emphasis on facial video editing, where preserving identity consistency becomes crucial. This is because inconsistencies in other categories may not lead to severe issues, but in the context of facial video, any inconsistency can lead to critical application failures. Notably, Kim \textit{et al.} \cite{kim2023dva} share a task closely aligned with our own, utilizing an identity encoder to extract identity features from the source video. In our method, we concentrate predominantly on ensuring consistency between the edited output and the source frame, achieving improved identity consistency without reliance on video data.

\subsection{Diffusion-based generative models.}
Denoising diffusion probabilistic models (DDPMs) \cite{ho2020denoising} are employed to link image generation with the sequential denoising of isotropic Gaussian noise. The diffusion model undergoes training to forecast the noise originating from the input image. The Latent Diffusion Model (LDM) \cite{rombach2022high} represents a subclass of DDPMs that are trained on the latent space. Recent advancements in diffusion models \cite{sohl2015deep} have facilitated cutting-edge image synthesis \cite{dhariwal2021beatgan,ho2022cascaded,saharia2022img2img,saharia2022superres,song2019generative}, and the latest text-to-image diffusion models \cite{nichol2021glide,ramesh2022hierarchical,saharia2022photorealistic} have demonstrated the capability to produce realistic images based on arbitrary text captions. In this study, we adopt the Latent Diffusion Model as the foundational framework for encoding, manipulating, and decoding real facial videos.

\subsection{Image editing with diffusion models.}
Given the potent image generation capabilities of diffusion models, numerous endeavors have been made to harness these models for image manipulation tasks, including text-guided inpainting \cite{nichol2021glide,avrahami2022blended,lugmayr2022repaint, couairon2022diffedit}, single image editing \cite{zhang2023sine}, and instruction-based image editing \cite{brooks2023instructpix2pix,yue2023chatface}. Among these, our particular focus lies in preserving the fidelity of the source image. Brooks~\cite{brooks2023instructpix2pix} introduce a method for text-based image manipulation by introducing an additional input channel to the UNet architecture within the diffusion model. Similarly, IP-Adapter~\cite{ye2023ip} also applies image features in diffusion models, but focuses mainly on the subject instead of consistency. In parallel, an alternative approach to image manipulation has been explored, known as the Diffusion Autoencoder (DiffAE)~\cite{preechakul2022diffae}, which employs a learnable encoder to extract semantic representations that condition the underlying diffusion model. However, when addressing the challenge of video manipulation, it has been observed that the work of Brooks \textit{et al.} \cite{brooks2023instructpix2pix} struggles to maintain subject consistency across frames, while DiffAE exhibits limitations in its capacity to manipulate a broad range of categories. To overcome these shortcomings, we propose a hybrid framework that integrates both text and image controls to enhance consistency. Furthermore, we leverage the capabilities of CLIP \cite{gal2022clipstylegan}, an essential tool for multi-modal tasks.

\subsection{Training data generation.}
Deep learning models typically demand substantial volumes of training data. In the context of facial video editing, numerous facial image datasets are available; however, the corresponding edited dataset is notably scarce. Previous studies \cite{li2022bigdatasetgan,ravuri2019classification,shrivastava2017learning,tritrong2021repurposing,brooks2023instructpix2pix} have employed techniques such as generative models and automated scripts to create extensive training data for subsequent tasks. In our framework, we adopt a professional rendering engine in conjunction with an image caption model \cite{li2022blip} to construct our paired dataset.

\section{Method}
Our goal is to generate an edited video from an authentic facial video and a target prompt, maintaining the original video's identity and ensuring seamless frame-to-frame consistency.

We began by developing a training dataset that includes source images, modified images of specific subjects, and associated captions for the edited images (Section~\ref{sec3.1}). Following this, we harnessed two diffusion models, trained on our dataset, to exercise precise control over text and image components. These models were subsequently utilized during the inference phase (Section~\ref{sec3.2}). To conclude the process, we implemented optimization strategies on the resultant video, focusing on enhancing its quality while ensuring consistent portrayal of the edited subject and identity (Section~\ref{sec3.3}).

\subsection{Generating Training Dataset}
\label{sec3.1}
The foundation of our fine-tuning process is dual modal guidance: images paired with text. Each data of the dataset comprises a source image, its edited counterpart, and an apt caption for the edited image.
Jiang \textit{et al.}~\cite{jiang2023identity} have underscored that when modifying facial content in videos, it's essential to ensure uniform editing effects across successive frames and from various angles and actions. Consequently, we curated facial images from multiple perspectives as our source images, sourcing them from several open-access datasets~\cite{schuhmann2021laion,karras2017progressive}. The first step involved creating the edited image from the source facial image, after which image caption models were used to produce the prompt.

\textbf{Generating edited images. }The process of generating edited images involves the utilization of a commercial software that is specifically used to create special effects for applying special effects to images centered on a specific subject. This procedure begins with the preparation of graphical assets for the desired effect, which we refer to as ``magic face'', encompassing makeup, filters, and props. Subsequently, we import both the source facial image and the material package into the rendering software. The software autonomously identifies key facial landmarks and appropriately applies the graphical materials to the corresponding positions on the face. This iterative process is applied to all source images, resulting in the creation of edited images. Our selected source image dataset encompasses faces of varying ages, genders, and perspectives, thereby enhancing the relative alignment of the edited subject with facial key landmarks. This heightened alignment contributes to maintaining consistency when generating video content.

\textbf{Generating prompts. }Textual guidance has been frequently adopted in text-to-image diffusion models~\cite{rombach2022high}. With our edited images at hand, the next step is to curate descriptive captions for them. We leverage the capabilities of the pre-trained BLIP model~\cite{li2022blip} to transmute edited images into prompts, encapsulating the core features of the source image. Concurrently, we aspire to build a text control model, inspired by the Dreambooth methodology \cite{ruiz2023dreambooth}. We attribute a unique identifier to each edited subject and synchronize it with a caption prompt. In this context, the ``magic face'' label serves as an apt identifier. The resultant dataset prompts amalgamate the BLIP model's output (``[caption]''), a conjunction (``[conj]''), and the unique ``magic face'' effect name (``[magic\_face\_name]''). This structured approach ensures the generated text encompasses both the essence of the edited image and a distinct subject signature.

\subsection{Training and models}
\label{sec3.2}
Diffusion models are a subset of probabilistic generative models crafted to understand data distributions by progressively denoising variables sampled from a Gaussian distribution. In our research, we particularly emphasize latent diffusion models (LDMs) \cite{rombach2022high} that have been trained on extensive datasets, forming the cornerstone of our methodology.

Our architecture encompasses two LDMs: one devoted to textual control and the other to image-based control. The text control model seeks to bridge the edit prompt with specific subjects like glasses, makeup, or other special effects. Conversely, the image control model is designed to ensure that the modified image mirrors the original facial identity, while also upholding consistency through successive frames.

\subsubsection{Subject-to-Text Control Integration}

The text control model we employ is essentially a text-to-image diffusion model \cite{rombach2022high} that has been refined using our distinctive dataset. The encoding process is overseen by the locked-weight vae encoder $E$ which converts a given image $x$ into its latent representation, $z$, such that $z = \mathcal{E}(x)$. Starting with an initial noise map, represented as $\epsilon \sim \mathcal{N}(0,1)$, the latent code can be formulated as $\mathbf{z_t} = \alpha_t \mathbf{z}+\sigma_t \boldsymbol{\epsilon}$. Taking cues from the text-to-subject integration methodology introduced in Dreambooth \cite{ruiz2023dreambooth}, we apply a comparable strategy to our video creation task. However, our focus is centered on maintaining the consistency of the edited subject, prioritizing the retention of the edited subject from our training set over preserving prior knowledge of the pre-trained model. This led to the exclusion of the prior-preservation term from our loss function formula. We craft the conditioning vector, symbolized as $\mathbf{c}$, by employing a text encoder in tandem with a prompt. The squared error loss during the training phase can be expressed as:

\begin{equation}
\mathbb{E}_{\mathcal{E}(x), \mathbf{c}, \boldsymbol{\epsilon}, t}\left[w_t\left\|\hat{\mathbf{\epsilon}}_\theta\left(\alpha_t \mathbf{z}+\sigma_t \boldsymbol{\epsilon}, t,\mathbf{c}\right)-\mathbf{\epsilon}\right\|_2^2\right],   
\end{equation}
where $\mathbf{z}$ denotes the latent of the ground-truth image, $\hat{\mathbf{\epsilon}}$ is the noise prediction model, and $\alpha_t, \sigma_t, w_t$ are variables that modulate the noise schedule and sample quality.

Interestingly, the core subject binding refinement method requires a minimal set of images, and our technique does not necessitate any extra images. This is attributed to the fact that the text control model is mainly used to subject consistency, which is to ensure a same edited subject among frames in the result. On the other hand, the principal oversight of video consistency is predominantly facilitated by the image control model.

\subsubsection{Additional Channel for Image Control}

The image control model is pivotal in safeguarding the integral attributes of the source image, thereby cementing the identity consistency of our approach. To infuse the information of the source image into our pre-trained model, we gleaned insights from InstructPix2Pix \cite{brooks2023instructpix2pix}, which suggests appending an extra input channel to the UNet framework. Specifically, for a given source image $x$, we encode it into the latent space and introduce it as an auxiliary input channel, symbolized as $\mathcal{E}(x)$, which is then concatenated with the noisy latent $\mathbf{z_t}$. Here, $\mathcal{E}$ represents the encoder of the latent diffusion. We initialize the weights of this newly integrated channel to zero, whereas the remaining weights are initialized based on the pre-trained model. The loss function to be minimized is articulated as:
\begin{equation}
\mathbb{E}_{\mathcal{E}(x), \epsilon \sim \mathcal{N}(0,1), t}\left[\| \epsilon-\epsilon_\theta\left(z_t, t, \mathcal{E}\left(x\right) \right) \|_2^2\right].   
\end{equation}

In the practical implementation of our model, it's noteworthy to mention that we consistently retain the text condition $c_T$ during training. However, during the referencing phase, we omit the noise prediction channel linked with the text condition, focusing only on the unconditional prediction and the image condition prediction. With a guiding scale of $s>=1$, the noise prediction for this model is determined as:
\begin{equation}
\tilde{e_I}\left(z_t, c_I\right)=e_I\left(z_t, \varnothing\right)+s \cdot\left(e_I\left(z_t, c_I\right)-e_I\left(z_t, \varnothing\right)\right),
\end{equation}
where $e_I(z_t, c_I)$ represents the image control network and $c_I$ is the latent of the input image.

\subsubsection{Integrated Guidance during Inference}

\begin{figure}[tp]
    \centering
    \includegraphics[width=\linewidth]{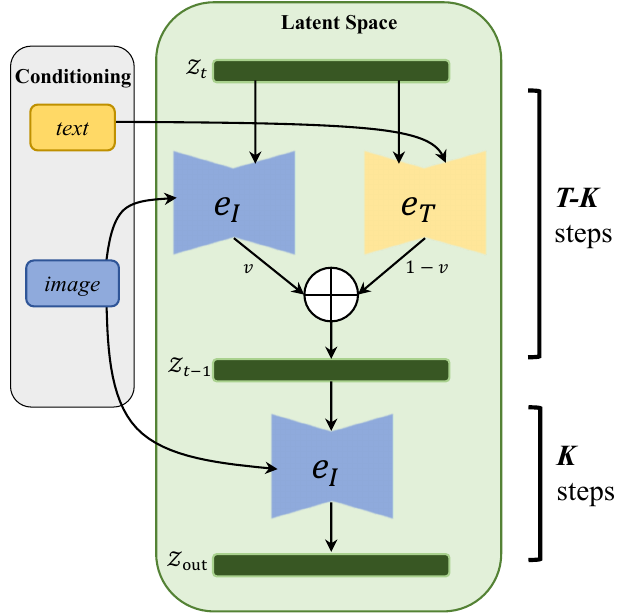}
    \caption{\textbf{Inference pipeline. }DiffMagicFace consists of two fine-tuned model based on latent diffusion models. During training, we fine-tune a text-control model and a image-control model. When sampling, the text-control model take the target text and a noisy latent code $z_t$ as input and the image-control model take a latent of source image and $z_t$ as input. Their noise prediction are made element-wise addition under weight $v$ for $T-K$ steps, and inference only on image-control model for $K$ steps. }
    \label{fig:inference pipeline}

\end{figure}

Figure \ref{fig:inference pipeline} illustrates the process of the inference phase, during which both the text and image control models operate in tandem. The process begins by deploying the latent code, represented as $z_T$, tailored to the spatial dimensions of the anticipated output. This code, $z_T$, in conjunction with the target prompt (capturing the desired effects), is then input into the text control model. In parallel, the latent code corresponding to the source image and $z_T$ is fed into the image control model. Here, we employ notations $e_T(z_t, c_T)$ and $e_I(z_t, c_I)$ to symbolize the text and image control models respectively. The amalgamated noise prediction for our dual-model system, denoted by $e_\theta(z_t, c_T, c_I)$, is derived from:

\begin{equation}
\tilde{e_\theta}\left(z_t, c_T, c_I\right)=
v \cdot e_T\left(z_t, c_T\right)+
(1-v)\cdot e_I\left(z_t, c_I\right).
\end{equation}

It is important to underscore that the image control model undergoes extensive training on a plenty of images spanning multiple angles and nuances of a particular subject. This extensive training ensures that it embodies a comprehensive understanding of the subject's visual subtleties. In contrast, the text control model has no need of this depth of training. 
From pre-experiments, we found that there is a strong correlation between the reality of the generated edited images and the identity consistency, while the identity consistency is mainly determined by the image control model. As for diffusion models, producing outputs of high visual fidelity is mainly affected by later sampling steps~\cite{balaji2022ediff}.
Hence, to maintain consistency and ensure the integrity of the final output, we predominantly rely on the image control model during the concluding steps of inference. Specifically, for steps where $t>K$, the latent code is derived using the above equation. However, for steps within the range of $0<t\leq K$, the denoising is exclusively entrusted to the image control model. Here, the hyperparameter $K$ plays a pivotal role in striking a delicate equilibrium between ensuring subject clarity and upholding the consistency of the edits.

\subsection{Video Consistency Optimization}
\label{sec3.3}

In this section, we introduce our approach for generating an edited video using the models and apply optimization techniques to enhance video consistency. By providing each frame of the source video and the target prompt to the model, the edited video is generated sequentially, frame by frame, with the same prompt and a fixed seed. Thanks to the training on a multi-perspective facial dataset and the image control channel, the generated video exhibits coherence in the edited subject (e.g. glasses moving in sync with the head's movement), and preserves the facial identity of the source video. 

To further improve the consistency and coherence within our videos, we have employed the prowess of two primary optimization techniques: optical flow and the low-pass filter.

\textbf{Optical flow. }Optical flow~\cite{beauchemin1995opticalflow} is a computer vision technique that estimates the motion of objects in image sequences or video frames and calculates the displacement of each pixel between consecutive frames. In our work, we apply optical flow to reduce video jitter by estimating motion vectors and stabilizing the video's motion through object motion compensation. We compute dense optical flow vectors~\cite{alvarez2000denseopticalflow} and apply inverse transformations to pixels to compensate for video content deformations.

\textbf{Low-pass filter. }A low-pass filter \cite{kejriwal2016lowpass} is a signal processing filter that allows low-frequency components to pass through while attenuating or blocking high-frequency components. To remove flicker in a video, a low-pass filter can be applied to reduce the high-frequency temporal variations in the video frames, effectively smoothing out rapid brightness or color fluctuations, which are typically responsible for inducing flicker.  This is achieved by averaging pixel values temporally,  resulting in a more consistent and flicker-free video.

With the incorporation of these optimization techniques, our edited videos  not only manifest enhanced frame-to-frame coherence but also register superior metrics in identity consistency, as shown in Table~\ref{table:abl_opt}.

\section{Experiments}

\label{sec:experiments}
\subsection{Implementation Details}

Our video editing method does not rely on specific video datasets. Instead, we selected $30,000$ human facial images from the CelebA-HQ dataset \cite{karras2017progressive}. These images were processed for eight special effect subjects using our dataset generation pipeline, resulting in a total of $240,000$ edited images. All images were resized to a resolution of $256 \times 256$ and subjected to random horizontal flipping. We fine-tuned our method using the text-to-image model LDM, specifically Stable Diffusion~\cite{rombach2022high}, which was originally trained on the LAION dataset~\cite{schuhmann2021laion}. For fine-tuning, we worked with a latent code spatial size of $32 \times 32$.

The text control model was fine-tuned by randomly selecting ten images for each subject. Our experiments were conducted using a single RTX 3090 GPU with a batch size of $1$ and $400$ training steps, with a base learning rate set to $5 \times 10^{-5}$. The image control model was fine-tuned on the full dataset, using one RTX 3090 GPU with a batch size of $4$ and $30,000$ training steps, and a learning rate of $5 \times 10^{-5}$.

For sampling parameters, we used inference steps $T=50$, $K=20$, and $v=0.9$. In the context of video manipulation, we applied the low-pass filter optimization twice, in addition to both optical flow and dense optical flow processes.

\subsection{Qualitative Evaluation}

\begin{figure*}[htbp]
\centering
\begin{tabular}{@{}p{0.03\textwidth}@{}p{0.95\textwidth}@{}}

\rotatebox{90}{Origin} & \includegraphics[width=\linewidth]{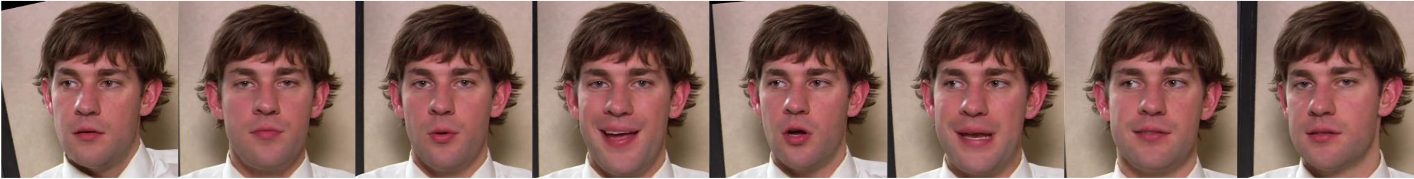} \\ 

\rotatebox{90}{IP2P~\cite{brooks2023instructpix2pix}}  & \includegraphics[width=\linewidth]{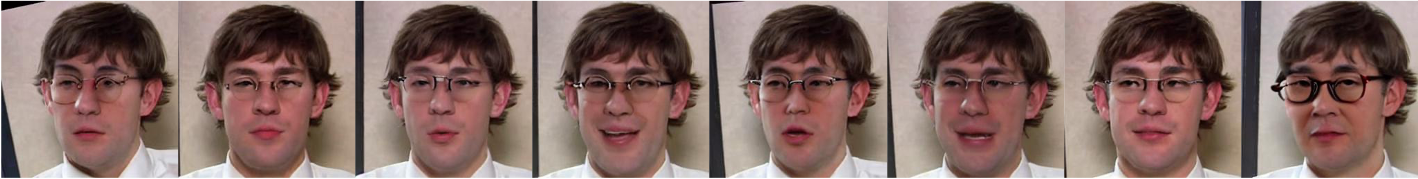} \\

\rotatebox{90}{DiffEdit~\cite{couairon2022diffedit}}  & \includegraphics[width=\linewidth]{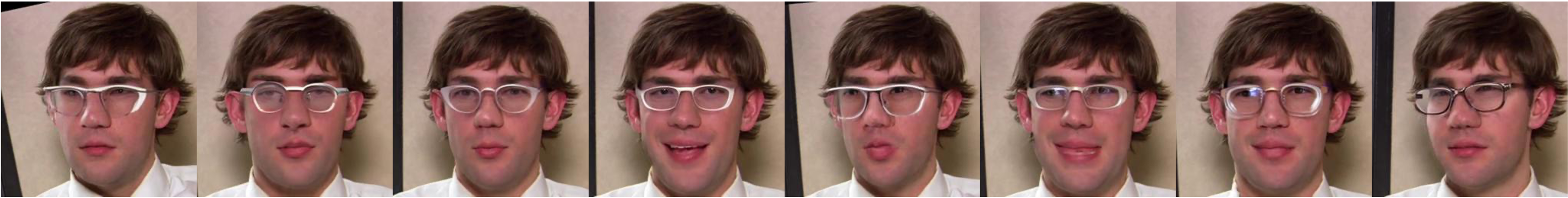} \\

\rotatebox{90}{DVA~\cite{kim2023dva}}  & \includegraphics[width=\linewidth]{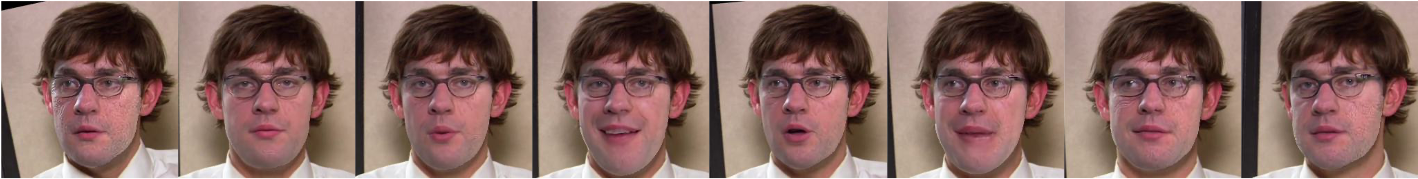} \\

\rotatebox{90}{Ours}  & \includegraphics[width=\linewidth]{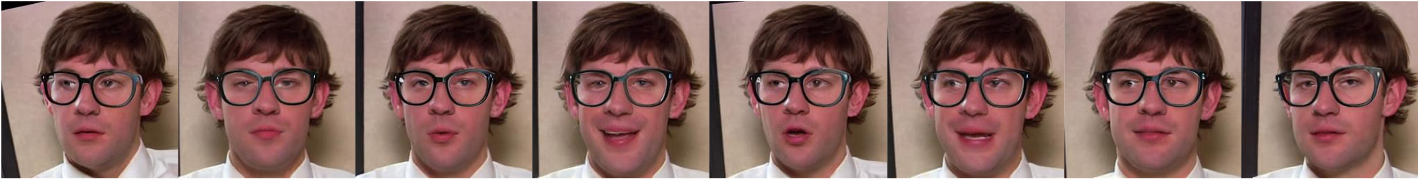} \\

\end{tabular}
\caption{Comparison of qualitative results with previous editing methods for the editing subject ``glasses''.}
\label{fig:comparison}
\end{figure*}

\begin{figure*}[htbp]
\centering
\begin{tabular}{@{}p{0.1\linewidth}@{}p{0.9\linewidth}@{}}

\begin{minipage}{0.1\textwidth} 
        \centering{Source Kamala}
\end{minipage}  & 
\begin{minipage}[c]{\textwidth} 
\includegraphics[width=0.9\linewidth]{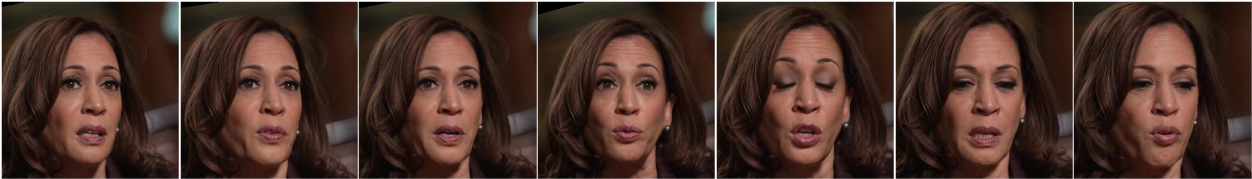} 
\end{minipage} \\

\begin{minipage}{0.1\textwidth} 
        \centering{\textit{a person in \textbf{sweet makeup}}}
\end{minipage}  & 
\begin{minipage}[c]{\textwidth} 
\includegraphics[width=0.9\linewidth]{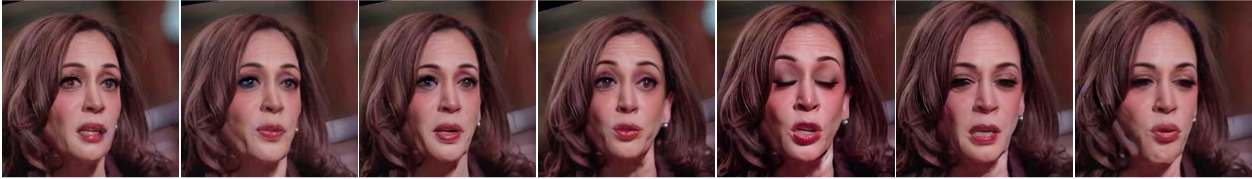} 
\end{minipage} \\

\begin{minipage}{0.1\textwidth} 
        \centering{\textit{a person wearing the \textbf{creamy facial mask}}}
\end{minipage}  & 
\begin{minipage}[c]{\textwidth} 
\includegraphics[width=0.9\linewidth]{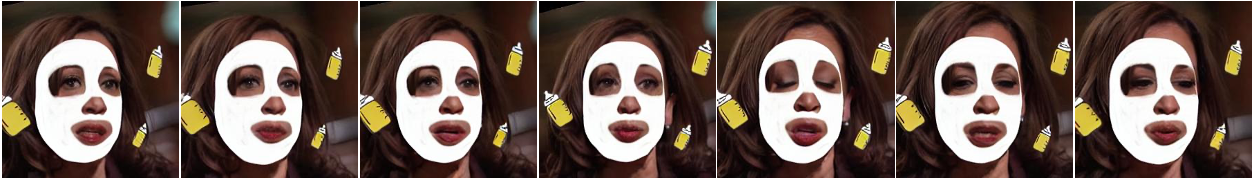} 
\end{minipage} \\

\begin{minipage}{0.1\textwidth} 
        \centering{Source Obama}
\end{minipage}  & 
\begin{minipage}[c]{\textwidth} 
\includegraphics[width=0.9\linewidth]{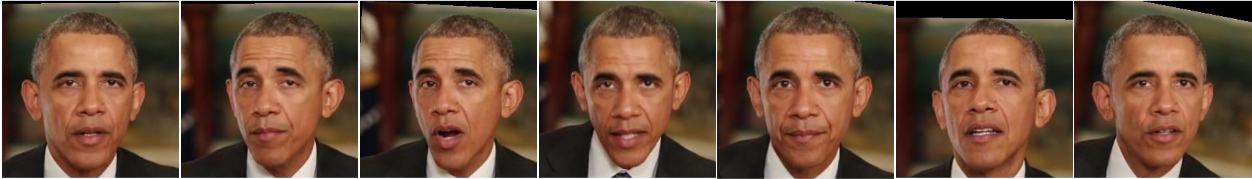} 
\end{minipage} \\

\begin{minipage}{0.1\textwidth} 
        \centering{\textit{a person with \textbf{green bow effect}}}
\end{minipage}  & 
\begin{minipage}[c]{\textwidth} 
\includegraphics[width=0.9\linewidth]{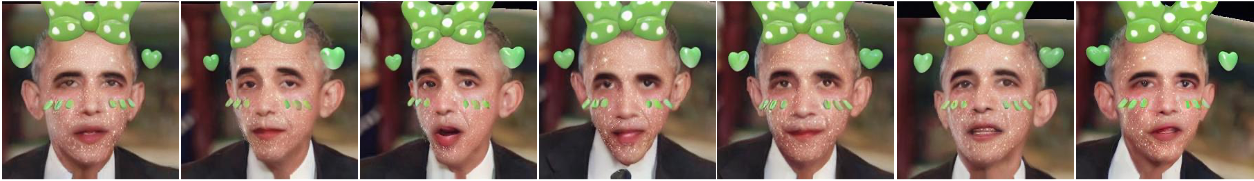} 
\end{minipage} \\

\begin{minipage}{0.1\textwidth} 
        \centering{\textit{a person with the \textbf{burning fire effect}}}
\end{minipage}  & 
\begin{minipage}[c]{\textwidth} 
\includegraphics[width=0.9\linewidth]{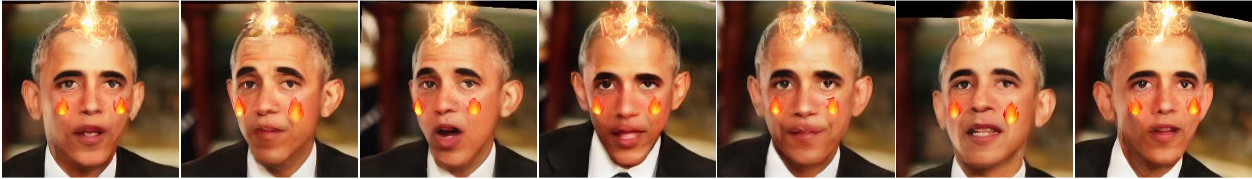} 
\end{minipage} \\

\end{tabular}
\caption{Displayed the visual effects of different types of special effects applied in our method. Based on different text inputs, our method can edit highly consistent special effects videos according to the corresponding subjects.}

\label{fig:visual_res}
\end{figure*}

In this section, we conduct a comprehensive qualitative evaluation of our DiffMagicFace video editing approach. We compare its performance visually with other existing methods, showcasing its ability to produce high-quality results.

In Figure \ref{fig:comparison}, we present a visual comparison of DiffMagicFace with other methods. We edited a facial video with these methods frame by frame to add a pair of glasses, guided by appropriate text prompts for each method. The results vividly illustrate the effectiveness of DiffMagicFace in generating realistic glasses while maintaining subject consistency between frames and preserving facial identity. In contrast, IP2P~\cite{brooks2023instructpix2pix}, which is a symbolic image editing method, struggles to maintain consistent rendering of glasses (the fifth frame and the last one), and also destroys part of the facial details (the last frame). DiffEdit~\cite{couairon2022diffedit} is a method that based on inpainting with generated masks, however, fails to maintain content consistency and human identity in this task. Inpainting can generate realistic for a single frame from the content out of the mask, but has no contribution to previous and next frames (Difference of glasses and eyes between the first and fourth frame). Although Diffusion Video Autoencoder~\cite{kim2023dva}, which is an video editing method focus on identity preservation, demonstrates improved consistency with the effect of its identity encoder in the method, its generated frames exhibit inconsistency in other parts of the face and lead to unrelated visual result (as seen in variations in skin color and eye contours). As for the result generated by our method, the frames not only possess the coherent editing subject since the color and shape detail of all the frames are same, but also perform best in maintaining facial identity, which can be observed by the location of eyeball in each frame, and the whole generated video is realistic and time-consistent.

Moving beyond basic facial props like glasses, we explore a wide range of ``magic-face'' effects in our experiments. As depicted in Figure \ref{fig:visual_res}, we apply two distinct magical facial effects, ``sweet makeup'' and ``creamy facial mask,'' specifically designed for the female subject in the source video. These effects not only include fundamental graphical elements but also incorporate filters, transformations, and brightness adjustments, representing abstract aspects within our dataset. The results underscore our model's ability to adeptly handle these features.

In a second set of experiments, we introduce two additional magical facial effects, labeled ``green bow effect" and ``burning fire effect." These effects primarily consist of picture stickers and animated elements that are position-sensitive, requiring precise placement on the face. The results demonstrate our method's capacity to learn and effectively apply these spatial relationships from the dataset during video inference. In summary, our approach seamlessly adapts to a diverse array of magical facial effects, producing visual results that closely resemble those achieved through rendering-based and hand-crafted special effects.

\begin{figure}[htp]
    \centering
    \includegraphics[width=\linewidth]{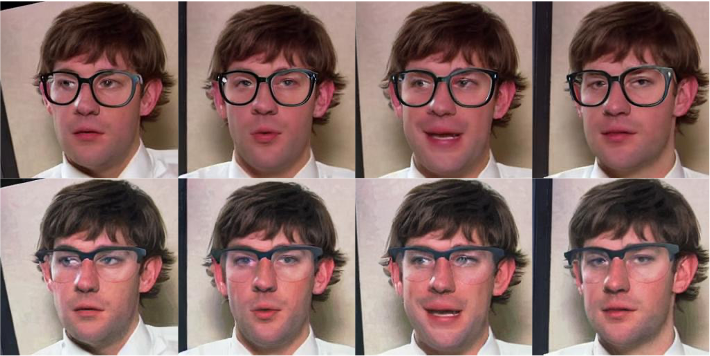}
    \caption{Distinguishing between the editing of similar subjects. Text guidances are ``a person wearing black full-frame glasses" (up) and ``a person wearing half-frame glasses" (below).}
    \label{fig:glasses_dis}

\end{figure}

Furthermore, Figure \ref{fig:glasses_dis} showcases our method's ability to distinguish between similar subjects. By employing various prompts containing the term ``glasses," we can generate distinct target results while maintaining clear semantics and consistency. In our dataset, we have two edited subjects related to glasses, namely ``half-frame glasses" and ``full black frame glasses." The results clearly demonstrate successful editing of both videos. Unlike previous methods that could only generate an average representation of ``glasses" based on text-based guidance, our framework incorporates an independent text control model specifically designed to differentiate between different subjects within our dataset.

\subsection{Quantitative Evaluation}

In this subsection, we perform a quantitative assessment of our DiffMagicFace video editing method. We aim to evaluate the temporal consistency of the edited videos presented in Figure \ref{fig:comparison}. It's worth noting that, as acknowledged by Kim \textit{et al.} \cite{kim2023dva}, there is currently no established standard metric for evaluating the temporal consistency of videos in the context of text-driven editing. Nevertheless, we employ two metrics, TL-ID (Temporal Local Identity Distance) and TG-ID (Temporal Global Identity Distance) \cite{tzaban2022stitch}, which serve as proxies to gauge both local and global consistency of identity across frames when compared to the original content. These metrics are of particular relevance to our task, with TL-ID specifically assessing the consistency of the edited subject's content.

In Table \ref{table:comp}, we provide the quantitative results of our method, DiffMagicFace, alongside other facial editing methods, Brooks \textit{et al.} \cite{brooks2023instructpix2pix} and Kim \textit{et al.} \cite{kim2023dva}. Notably, our task involves editing glasses onto a face, a more challenging endeavor compared to previous works. It is evident from the table that DiffMagicFace achieves the highest TL-ID and TG-ID scores among the three methods, indicating superior temporal consistency and identity preservation in the edited videos.

\begin{table}
  \centering
  \caption{\textbf{Quantitative results} to evaluate the consistency of the edited video.}
  \label{table:comp}
  \begin{tabular}{l|cc}
  \toprule
    Method & TL-ID & TG-ID \\
    \midrule
    InstructPix2Pix~\cite{brooks2023instructpix2pix} & 0.646 & 0.642 \\
    DiffEdit~\cite{couairon2022diffedit} & 0.760 & 0.708 \\
    Diffusion Video Autoencoder~\cite{kim2023dva} & 0.901 & 0.857 \\
    Ours & \textbf{0.993} & \textbf{0.912} \\
    \bottomrule
  \end{tabular}
\end{table}

To further explain the application value of our method, we conduct an experiment of execution time among the rendering software and all compared methods. Since each method has a different pre-loading process, for the sake of fairness, we only calculate and compare the time of single-frame editing image generation in our experiments. Specifically, for InstructPix2Pix~\cite{brooks2023instructpix2pix}, DiffEdit~\cite{couairon2022diffedit} and our method DiffMagicFace, we exclude the interference of initialization time and only calculate the average time taken to reason through the diffusion process for 100 frames of images in the video shown in Figure \ref{fig:comparison}. For Diffusion Video AutoEncoder~\cite{kim2023dva}, their method needs to temporarily train the identity encoder on the input video during each inference, so the cost of training needs to be added to the total time cost of the inference process as a part of the average calculation. For the specific rendering software, we queried the standard interface of the engine locally in this experiment, performed special effects rendering frame by frame, output the video results and obtained the average single frame time. Note that all these model-based method is running with 1 RTX 3090 GPU and set 30 diffusion inference steps in this experiment.

\begin{table}
  \centering
  \caption{Comparison of the per-frame execution time cost of each method.}
  \label{table:exe_time}
  \begin{tabular}{l|cc}
  \toprule
    Method & Time Cost (second/frame)\\
    \midrule
    Rendering Software~\cite{be} & 4.6326 \\
    InstructPix2Pix~\cite{brooks2023instructpix2pix} & \textbf{0.9357} \\
    DiffEdit~\cite{couairon2022diffedit} & 15.9757 \\
    Diffusion Video Autoencoder~\cite{kim2023dva} & 103.6842 \\
    Ours & \textbf{2.3572}  \\
    \bottomrule
  \end{tabular}
\end{table}

Table \ref{table:exe_time} shows the time cost of each method during the inference step of the video editing task of Figure \ref{fig:comparison}. Our method exhibits relatively good inference performance and cost less time than that of the rendering software, which is used in our dataset creation. Considering the time-consuming advantage of inference, our method is more conducive to extending to some practical applications than rendering softwares, such as live broadcast special effects. In addition, our method is model-based and can be compatible with dozens or hundreds of special effects through fine-tuning training. Compared with the rendering engine using complex material packages, our method has obvious expansion and application value. As for other methods, the times cost result is consistent with the complexity of their models and frameworks. InstructPix2Pix~\cite{brooks2023instructpix2pix} is the most closest method to the plain image-to-image inference, thus has the lowest cost. Our method is combined with image-control model and a text-control model, so the inference time is about double of the lowest one, but also acceptable for applications. DiffEdit~\cite{couairon2022diffedit} is a method based on inpainting and its extra cost comes from the generation of the mask image, which leads to the explosion of time. Diffusion Video Autoencoder~\cite{kim2023dva} exhibits the most average inference time cost because training the identity encoder for the source video is a necessary part of their framework.

\subsection{Ablation Study}

\begin{table}
  \centering
  \caption{Ablation of the text control model $e_T$. Average scores on three edited subjects to substantiate the contribution to the consistency of edited video.}
  \label{table:abl_txt}
  \begin{tabular}{l|cc}
  \toprule
    Pipeline & TL-ID & TG-ID \\
    \midrule
    w/o $e_T$& 0.927 & 0.890 \\
    Full & \textbf{0.936} & \textbf{0.901} \\
    \bottomrule
  \end{tabular}
\end{table}
In this section, we conduct an ablation study to assess the contributions of various components within our framework.

\paragraph{Impact of Text Control Model}
One key component in our framework is the text control model, denoted as $e_T$. It acts as a vital role in maintaining consistency between frames by incorporating textual prompts into the editing process. To validate the necessity of the text control model, we conducted a quantitative experiment comparing our hybrid models, which include the text control model, with a pipeline lacking an independent text control model, akin to the implementation of IP2P \cite{brooks2023instructpix2pix}. We selected three distinct edited subjects for this experiment: ``wearing glasses,'' ``wearing a white straw hat,'' and ``head fire effect.'' Subsequently, we computed the average identity consistency metrics of the framework without and with $e_T$. The results, presented in Table \ref{table:abl_txt}, affirm that our independent text control model provides a reliable intermediate result $Z_{t-k}$ from its contribution in early $T-K$ sampling steps, and undeniably enhances the consistency of the edited video.
Note that our task is image editing, which has to be performed on the basis of the source image. The input of source image information is essential, so the indispensability of image control model $e_I$ is apparent and we skipped this part's ablation.

\paragraph{Impact of Optimization Techniques}
We further evaluate the impact of the optimization techniques employed in our framework. These techniques include optical flow and low-pass filtering debouncing. To validate their effectiveness, we applied these optimizations independently and in combination to assess their impact on the edited videos. The results, as shown in Table \ref{table:abl_opt}, indicate that the combination of these optimization techniques yields the most substantial improvement in the video consistency metric. Notably, low-pass filtering debouncing demonstrates pronounced effectiveness, and we applied this specific optimization technique twice during the generation of multiple results.
\begin{table}
  \centering
  \caption{Ablation of the optimizations. To validate that these optimizations make sense to the edited video.}
  \label{table:abl_opt}
  \begin{tabular}{cc|cc}
  \toprule
    Optical Flow & Low-Pass & TL-ID & TG-ID \\
    \midrule
     & & 0.906 & 0.872 \\
    \checkmark & & 0.952 & 0.890 \\
    & \checkmark & 0.979 & 0.905 \\
    \checkmark & \checkmark & \textbf{0.993} & \textbf{0.912} \\
    \bottomrule
  \end{tabular}
\end{table}

\section{Conclusion}
\label{sec:conclusion}

In conclusion, we've presented an innovative framework for effect-style editing facial videos while preserving identity. Our approach excels in producing highly consistent and apparently editing semantics. The results of  \textit{DiffMagicFace} achieve better results compared with previous methods, both visually and quantitatively. By fine-tuning the text control model and integrating an image control model, we've ensured clear and consistent subject editing while preserving facial identity. Notably, our training process doesn't rely on video datasets but leverages a systematic pipeline involving rendering softwares and image captioning models to generate a dataset. We've also employed optimization techniques to effectively improve video consistency. 
We plan to make our code, dataset, and dataset creation scripts publicly available, contributing valuable resources to the research community.


Despite its strengths, \textit{DiffMagicFace} faces limitations, particularly in handling editing subjects not included in the training set. Future work will aim to address this challenge, focusing on enhancing the generalizability of our control models to accommodate a wider array of editing categories.

\clearpage
\bibliography{references}

@article{karras2017progressive,
  title={Progressive growing of gans for improved quality, stability, and variation},
  author={Karras, Tero and Aila, Timo and Laine, Samuli and Lehtinen, Jaakko},
  journal={arXiv preprint arXiv:1710.10196},
  year={2017}
}

@inproceedings{rombach2022high,
  title={High-resolution image synthesis with latent diffusion models},
  author={Rombach, Robin and Blattmann, Andreas and Lorenz, Dominik and Esser, Patrick and Ommer, Bj{\"o}rn},
  booktitle={Proceedings of the IEEE/CVF conference on computer vision and pattern recognition},
  pages={10684--10695},
  year={2022}
}

@article{schuhmann2021laion,
  title={Laion-400m: Open dataset of clip-filtered 400 million image-text pairs},
  author={Schuhmann, Christoph and Vencu, Richard and Beaumont, Romain and Kaczmarczyk, Robert and Mullis, Clayton and Katta, Aarush and Coombes, Theo and Jitsev, Jenia and Komatsuzaki, Aran},
  journal={arXiv preprint arXiv:2111.02114},
  year={2021}
}

@inproceedings{li2022blip,
  title={Blip: Bootstrapping language-image pre-training for unified vision-language understanding and generation},
  author={Li, Junnan and Li, Dongxu and Xiong, Caiming and Hoi, Steven},
  booktitle={International Conference on Machine Learning},
  pages={12888--12900},
  year={2022},
  organization={PMLR}
}

@inproceedings{ruiz2023dreambooth,
  title={Dreambooth: Fine tuning text-to-image diffusion models for subject-driven generation},
  author={Ruiz, Nataniel and Li, Yuanzhen and Jampani, Varun and Pritch, Yael and Rubinstein, Michael and Aberman, Kfir},
  booktitle={Proceedings of the IEEE/CVF Conference on Computer Vision and Pattern Recognition},
  pages={22500--22510},
  year={2023}
}

@inproceedings{brooks2023instructpix2pix,
  title={Instructpix2pix: Learning to follow image editing instructions},
  author={Brooks, Tim and Holynski, Aleksander and Efros, Alexei A},
  booktitle={Proceedings of the IEEE/CVF Conference on Computer Vision and Pattern Recognition},
  pages={18392--18402},
  year={2023}
}

@inproceedings{preechakul2022diffae,
  title={Diffusion autoencoders: Toward a meaningful and decodable representation},
  author={Preechakul, Konpat and Chatthee, Nattanat and Wizadwongsa, Suttisak and Suwajanakorn, Supasorn},
  booktitle={Proceedings of the IEEE/CVF Conference on Computer Vision and Pattern Recognition},
  pages={10619--10629},
  year={2022}
}

@inproceedings{kim2023dva,
  title={Diffusion Video Autoencoders: Toward Temporally Consistent Face Video Editing via Disentangled Video Encoding},
  author={Kim, Gyeongman and Shim, Hajin and Kim, Hyunsu and Choi, Yunjey and Kim, Junho and Yang, Eunho},
  booktitle={Proceedings of the IEEE/CVF Conference on Computer Vision and Pattern Recognition},
  pages={6091--6100},
  year={2023}
}

@article{beauchemin1995opticalflow,
  title={The computation of optical flow},
  author={Beauchemin, Steven S. and Barron, John L.},
  journal={ACM computing surveys (CSUR)},
  volume={27},
  number={3},
  pages={433--466},
  year={1995},
  publisher={ACM New York, NY, USA}
}

@article{alvarez2000denseopticalflow,
  title={Reliable estimation of dense optical flow fields with large displacements},
  author={Alvarez, Luis and Weickert, Joachim and S{\'a}nchez, Javier},
  journal={International Journal of Computer Vision},
  volume={39},
  pages={41--56},
  year={2000},
  publisher={Springer}
}

@article{kejriwal2016lowpass,
  title={A hybrid filtering approach of digital video stabilization for UAV using kalman and low pass filter},
  author={Kejriwal, Lakshya and Singh, Indu},
  journal={Procedia Computer Science},
  volume={93},
  pages={359--366},
  year={2016},
  publisher={Elsevier}
}

@inproceedings{tzaban2022stitch,
  title={Stitch it in time: Gan-based facial editing of real videos},
  author={Tzaban, Rotem and Mokady, Ron and Gal, Rinon and Bermano, Amit and Cohen-Or, Daniel},
  booktitle={SIGGRAPH Asia 2022 Conference Papers},
  pages={1--9},
  year={2022}
}

@article{ramesh2022dalle2,
  title={Hierarchical text-conditional image generation with clip latents},
  author={Ramesh, Aditya and Dhariwal, Prafulla and Nichol, Alex and Chu, Casey and Chen, Mark},
  journal={arXiv preprint arXiv:2204.06125},
  volume={1},
  number={2},
  pages={3},
  year={2022}
}

@article{saharia2022imagen,
  title={Photorealistic text-to-image diffusion models with deep language understanding},
  author={Saharia, Chitwan and Chan, William and Saxena, Saurabh and Li, Lala and Whang, Jay and Denton, Emily L and Ghasemipour, Kamyar and Gontijo Lopes, Raphael and Karagol Ayan, Burcu and Salimans, Tim and others},
  journal={Advances in Neural Information Processing Systems},
  volume={35},
  pages={36479--36494},
  year={2022}
}

@article{ho2020denoising,
  title={Denoising diffusion probabilistic models},
  author={Ho, Jonathan and Jain, Ajay and Abbeel, Pieter},
  journal={Advances in neural information processing systems},
  volume={33},
  pages={6840--6851},
  year={2020}
}

@inproceedings{sohl2015deep,
  title={Deep unsupervised learning using nonequilibrium thermodynamics},
  author={Sohl-Dickstein, Jascha and Weiss, Eric and Maheswaranathan, Niru and Ganguli, Surya},
  booktitle={International conference on machine learning},
  pages={2256--2265},
  year={2015},
  organization={PMLR}
}

@article{dhariwal2021beatgan,
  title={Diffusion models beat gans on image synthesis},
  author={Dhariwal, Prafulla and Nichol, Alexander},
  journal={Advances in neural information processing systems},
  volume={34},
  pages={8780--8794},
  year={2021}
}

@article{ho2022cascaded,
  title={Cascaded diffusion models for high fidelity image generation},
  author={Ho, Jonathan and Saharia, Chitwan and Chan, William and Fleet, David J and Norouzi, Mohammad and Salimans, Tim},
  journal={The Journal of Machine Learning Research},
  volume={23},
  number={1},
  pages={2249--2281},
  year={2022},
  publisher={JMLRORG}
}

@inproceedings{saharia2022img2img,
  title={Palette: Image-to-image diffusion models},
  author={Saharia, Chitwan and Chan, William and Chang, Huiwen and Lee, Chris and Ho, Jonathan and Salimans, Tim and Fleet, David and Norouzi, Mohammad},
  booktitle={ACM SIGGRAPH 2022 Conference Proceedings},
  pages={1--10},
  year={2022}
}

@article{saharia2022superres,
  title={Image super-resolution via iterative refinement},
  author={Saharia, Chitwan and Ho, Jonathan and Chan, William and Salimans, Tim and Fleet, David J and Norouzi, Mohammad},
  journal={IEEE Transactions on Pattern Analysis and Machine Intelligence},
  volume={45},
  number={4},
  pages={4713--4726},
  year={2022},
  publisher={IEEE}
}

@article{song2019generative,
  title={Generative modeling by estimating gradients of the data distribution},
  author={Song, Yang and Ermon, Stefano},
  journal={Advances in neural information processing systems},
  volume={32},
  year={2019}
}

@article{nichol2021glide,
  title={Glide: Towards photorealistic image generation and editing with text-guided diffusion models},
  author={Nichol, Alex and Dhariwal, Prafulla and Ramesh, Aditya and Shyam, Pranav and Mishkin, Pamela and McGrew, Bob and Sutskever, Ilya and Chen, Mark},
  journal={arXiv preprint arXiv:2112.10741},
  year={2021}
}

@article{ramesh2022hierarchical,
  title={Hierarchical text-conditional image generation with clip latents},
  author={Ramesh, Aditya and Dhariwal, Prafulla and Nichol, Alex and Chu, Casey and Chen, Mark},
  journal={arXiv preprint arXiv:2204.06125},
  volume={1},
  number={2},
  pages={3},
  year={2022}
}

@article{saharia2022photorealistic,
  title={Photorealistic text-to-image diffusion models with deep language understanding},
  author={Saharia, Chitwan and Chan, William and Saxena, Saurabh and Li, Lala and Whang, Jay and Denton, Emily L and Ghasemipour, Kamyar and Gontijo Lopes, Raphael and Karagol Ayan, Burcu and Salimans, Tim and others},
  journal={Advances in Neural Information Processing Systems},
  volume={35},
  pages={36479--36494},
  year={2022}
}

@inproceedings{avrahami2022blended,
  title={Blended diffusion for text-driven editing of natural images},
  author={Avrahami, Omri and Lischinski, Dani and Fried, Ohad},
  booktitle={Proceedings of the IEEE/CVF Conference on Computer Vision and Pattern Recognition},
  pages={18208--18218},
  year={2022}
}

@inproceedings{lugmayr2022repaint,
  title={Repaint: Inpainting using denoising diffusion probabilistic models},
  author={Lugmayr, Andreas and Danelljan, Martin and Romero, Andres and Yu, Fisher and Timofte, Radu and Van Gool, Luc},
  booktitle={Proceedings of the IEEE/CVF Conference on Computer Vision and Pattern Recognition},
  pages={11461--11471},
  year={2022}
}

@article{yue2023chatface,
  title={ChatFace: Chat-Guided Real Face Editing via Diffusion Latent Space Manipulation},
  author={Yue, Dongxu and Guo, Qin and Ning, Munan and Cui, Jiaxi and Zhu, Yuesheng and Yuan, Li},
  journal={arXiv preprint arXiv:2305.14742},
  year={2023}
}

@article{gal2022clipstylegan,
  title={StyleGAN-NADA: CLIP-guided domain adaptation of image generators},
  author={Gal, Rinon and Patashnik, Or and Maron, Haggai and Bermano, Amit H and Chechik, Gal and Cohen-Or, Daniel},
  journal={ACM Transactions on Graphics (TOG)},
  volume={41},
  number={4},
  pages={1--13},
  year={2022},
  publisher={ACM New York, NY, USA}
}

@inproceedings{zhang2023sine,
  title={Sine: Single image editing with text-to-image diffusion models},
  author={Zhang, Zhixing and Han, Ligong and Ghosh, Arnab and Metaxas, Dimitris N and Ren, Jian},
  booktitle={Proceedings of the IEEE/CVF Conference on Computer Vision and Pattern Recognition},
  pages={6027--6037},
  year={2023}
}

@article{jiang2023identity,
  title={Identity-Aware and Shape-Aware Propagation of Face Editing in Videos},
  author={Jiang, Yue-Ren and Chen, Shu-Yu and Fu, Hongbo and Gao, Lin},
  journal={IEEE Transactions on Visualization and Computer Graphics},
  year={2023},
  publisher={IEEE}
}

@article{abdal2022video2stylegan,
  title={Video2stylegan: Disentangling local and global variations in a video},
  author={Abdal, Rameen and Zhu, Peihao and Mitra, Niloy J and Wonka, Peter},
  journal={arXiv preprint arXiv:2205.13996},
  year={2022}
}

@inproceedings{skorokhodov2022styleganv,
  title={Stylegan-v: A continuous video generator with the price, image quality and perks of stylegan2},
  author={Skorokhodov, Ivan and Tulyakov, Sergey and Elhoseiny, Mohamed},
  booktitle={Proceedings of the IEEE/CVF Conference on Computer Vision and Pattern Recognition},
  pages={3626--3636},
  year={2022}
}

@inproceedings{yao2021latent,
  title={A latent transformer for disentangled face editing in images and videos},
  author={Yao, Xu and Newson, Alasdair and Gousseau, Yann and Hellier, Pierre},
  booktitle={Proceedings of the IEEE/CVF international conference on computer vision},
  pages={13789--13798},
  year={2021}
}

@article{molad2023dreamix,
  title={Dreamix: Video diffusion models are general video editors},
  author={Molad, Eyal and Horwitz, Eliahu and Valevski, Dani and Acha, Alex Rav and Matias, Yossi and Pritch, Yael and Leviathan, Yaniv and Hoshen, Yedid},
  journal={arXiv preprint arXiv:2302.01329},
  year={2023}
}

@article{ho2022imagenvideo,
  title={Imagen video: High definition video generation with diffusion models},
  author={Ho, Jonathan and Chan, William and Saharia, Chitwan and Whang, Jay and Gao, Ruiqi and Gritsenko, Alexey and Kingma, Diederik P and Poole, Ben and Norouzi, Mohammad and Fleet, David J and others},
  journal={arXiv preprint arXiv:2210.02303},
  year={2022}
}

@article{singer2022makeavideo,
  title={Make-a-video: Text-to-video generation without text-video data},
  author={Singer, Uriel and Polyak, Adam and Hayes, Thomas and Yin, Xi and An, Jie and Zhang, Songyang and Hu, Qiyuan and Yang, Harry and Ashual, Oron and Gafni, Oran and others},
  journal={arXiv preprint arXiv:2209.14792},
  year={2022}
}

@inproceedings{ceylan2023pix2video,
  title={Pix2video: Video editing using image diffusion},
  author={Ceylan, Duygu and Huang, Chun-Hao P and Mitra, Niloy J},
  booktitle={Proceedings of the IEEE/CVF International Conference on Computer Vision},
  pages={23206--23217},
  year={2023}
}

@article{liu2023videop2p,
  title={Video-p2p: Video editing with cross-attention control},
  author={Liu, Shaoteng and Zhang, Yuechen and Li, Wenbo and Lin, Zhe and Jia, Jiaya},
  journal={arXiv preprint arXiv:2303.04761},
  year={2023}
}

@inproceedings{li2022bigdatasetgan,
  title={BigDatasetGAN: Synthesizing ImageNet with pixel-wise annotations},
  author={Li, Daiqing and Ling, Huan and Kim, Seung Wook and Kreis, Karsten and Fidler, Sanja and Torralba, Antonio},
  booktitle={Proceedings of the IEEE/CVF Conference on Computer Vision and Pattern Recognition},
  pages={21330--21340},
  year={2022}
}

@article{ravuri2019classification,
  title={Classification accuracy score for conditional generative models},
  author={Ravuri, Suman and Vinyals, Oriol},
  journal={Advances in neural information processing systems},
  volume={32},
  year={2019}
}

@inproceedings{shrivastava2017learning,
  title={Learning from simulated and unsupervised images through adversarial training},
  author={Shrivastava, Ashish and Pfister, Tomas and Tuzel, Oncel and Susskind, Joshua and Wang, Wenda and Webb, Russell},
  booktitle={Proceedings of the IEEE conference on computer vision and pattern recognition},
  pages={2107--2116},
  year={2017}
}

@inproceedings{tritrong2021repurposing,
  title={Repurposing gans for one-shot semantic part segmentation},
  author={Tritrong, Nontawat and Rewatbowornwong, Pitchaporn and Suwajanakorn, Supasorn},
  booktitle={Proceedings of the IEEE/CVF conference on computer vision and pattern recognition},
  pages={4475--4485},
  year={2021}
}

@article{couairon2022diffedit,
  title={Diffedit: Diffusion-based semantic image editing with mask guidance},
  author={Couairon, Guillaume and Verbeek, Jakob and Schwenk, Holger and Cord, Matthieu},
  journal={arXiv preprint arXiv:2210.11427},
  year={2022}
}

@article{balaji2022ediff,
  title={ediff-i: Text-to-image diffusion models with an ensemble of expert denoisers},
  author={Balaji, Yogesh and Nah, Seungjun and Huang, Xun and Vahdat, Arash and Song, Jiaming and Zhang, Qinsheng and Kreis, Karsten and Aittala, Miika and Aila, Timo and Laine, Samuli and others},
  journal={arXiv preprint arXiv:2211.01324},
  year={2022}
}

@article{gal2022textual,
  title={An image is worth one word: Personalizing text-to-image generation using textual inversion},
  author={Gal, Rinon and Alaluf, Yuval and Atzmon, Yuval and Patashnik, Or and Bermano, Amit H and Chechik, Gal and Cohen-Or, Daniel},
  journal={arXiv preprint arXiv:2208.01618},
  year={2022}
}

@inproceedings{kim2021exploiting,
  title={Exploiting spatial dimensions of latent in gan for real-time image editing},
  author={Kim, Hyunsu and Choi, Yunjey and Kim, Junho and Yoo, Sungjoo and Uh, Youngjung},
  booktitle={Proceedings of the IEEE/CVF Conference on Computer Vision and Pattern Recognition},
  pages={852--861},
  year={2021}
}

@inproceedings{cheng2020sequential,
  title={Sequential attention GAN for interactive image editing},
  author={Cheng, Yu and Gan, Zhe and Li, Yitong and Liu, Jingjing and Gao, Jianfeng},
  booktitle={Proceedings of the 28th ACM international conference on multimedia},
  pages={4383--4391},
  year={2020}
}

@inproceedings{wu2023tune,
  title={Tune-a-video: One-shot tuning of image diffusion models for text-to-video generation},
  author={Wu, Jay Zhangjie and Ge, Yixiao and Wang, Xintao and Lei, Stan Weixian and Gu, Yuchao and Shi, Yufei and Hsu, Wynne and Shan, Ying and Qie, Xiaohu and Shou, Mike Zheng},
  booktitle={Proceedings of the IEEE/CVF International Conference on Computer Vision},
  pages={7623--7633},
  year={2023}
}

@article{ye2023ip,
  title={Ip-adapter: Text compatible image prompt adapter for text-to-image diffusion models},
  author={Ye, Hu and Zhang, Jun and Liu, Sibo and Han, Xiao and Yang, Wei},
  journal={arXiv preprint arXiv:2308.06721},
  year={2023}
}

@misc{be,
  author       = "Kwai",
  title        = "Beyond Effects",
  howpublished = "https://effect.kuaishou.com/",
}
\bibliographystyle{IEEEtran}

\clearpage
\setcounter{figure}{0}
\setcounter{section}{0}
\section*{Supplementary Material}
\subsection{Our Dataset and Tools}

In the process of dataset creation, we employed the BeyondEffect~\cite{be} rendering engine, a commercial software typically utilized by professional special effects artists. When employing this software, it is imperative for the creators to prefabricate asset packages. As demonstrated in our paper, we have crafted over a dozen asset packages commonly used in short video production, each comprising various image files, filter files, and coordinate configuration files. We strategically positioned the special effects at specific relative locations on the key points of template faces. During rendering, the software automatically detects faces within the images and places the special effects according to the preset keypoint coordinates. 

\begin{figure}[htp]
    \centering
    \includegraphics[width=\linewidth]{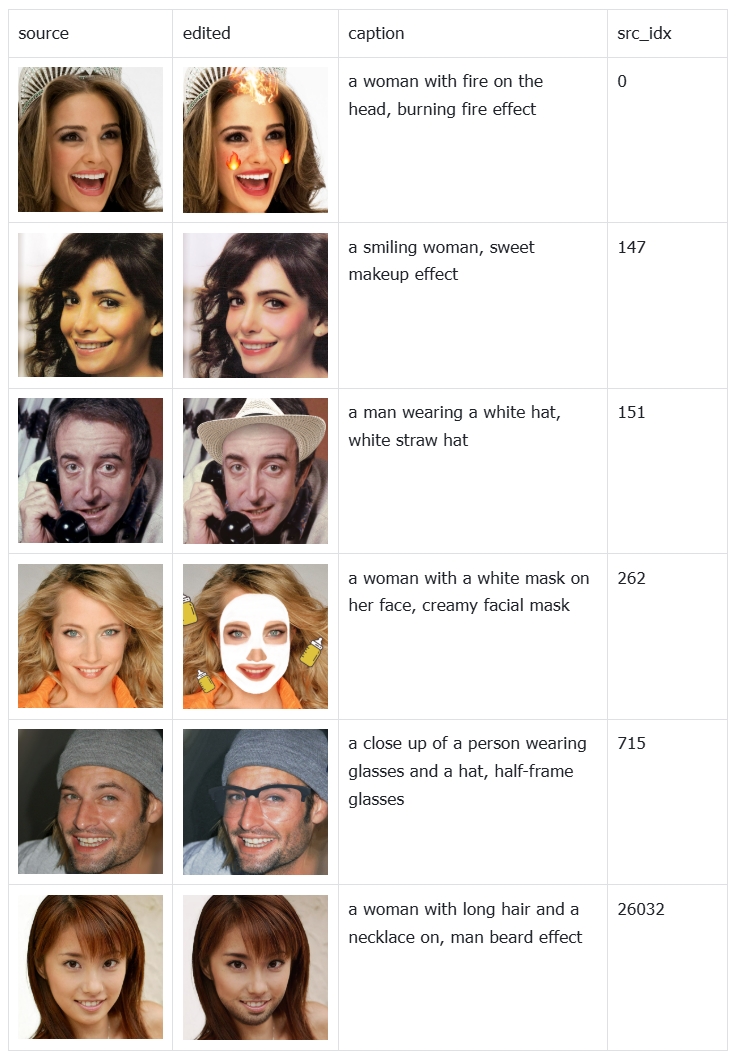}
    \caption{Samples of out dataset.}
    \label{fig:dataset}

\end{figure}

In this work, we have developed an automated script to invoke the interfaces provided by the software. Once the asset packages are imported and configured, the software can render the 30,000 facial images from the CelebaHQ dataset~\cite{karras2017progressive}, yielding edited results. The generation time for each special effects category dataset is approximately three days. We are willing to fully disclose the script for data generation and the special effects asset packages that have been created. Users who wish to generate custom datasets can collect their own asset packages and utilize our script to do so.

An example of the dataset we generated is shown in Figure~\ref{fig:dataset}, which includes the source image, the edited image, and the text. The text is obtained by performing image captioning on the edited image using BLIP2~\cite{li2022blip} and then concatenating the special effects name.

In the generated datasets, each category contains specific filters, such as the dark filters in the "man beard effect" category. These visual effects appear in the ground truth as part of the rendering results, making it reasonable for the final edited results produced by DiffMagicFace to also include the corresponding filters. This approach ensures that the datasets are not only visually rich but also contextually accurate, providing a robust foundation for further analysis and application in the field of special effects and computer vision.

Note that using this software is only one of the methods we use to create datasets. The hybrid control model we propose itself can learn specific editing subjects from a large number of editing data pairs and use the corresponding prompts during inference. In other words, our framework can model the special effects provided by the software, but is not limited to these special effects categories of the software. As for the difference between using our hybrid control model for video editing and rendering software for video editing, our model can accommodate more categories and is text-driven at inference time, while rendering software requires users to obtain specific material packages.

\subsection{Applications after DiffMagicFace}
\textit{DiffMagicFace} holds considerable practical potential, especially in scenarios where special effects are in demand. Traditionally, creating such effects required professional software such as BeyondEffect, and preparing intricate material packs. The whole process poses challenges for everyday users and short video platforms. 
Our framework offers a more accessible alternative by leveraging deep learning models, specifically diffusion models, to democratize the creation of special effects. After training a specific category of special effects with \textit{DiffMagicFace}, users can generate effects by simply specifying a brief text prompt, significantly reducing the operational complexity from the user's perspective.

Meanwhile, for creators of "magic-face" effects who have developed a limited set of effects (typically numbering in the tens) and aim to share them with their audience, an effective approach would be to train a private model, publish the corresponding user interface, and allow users to perform video editing through text-based prompts.

\subsection{Extra Comparison}
This manuscript addresses the task of video editing that preserves facial identity features. To our knowledge, all works that emphasize the consistency of identity are discussed in the Comparison section of the main text. To elucidate the task more clearly, this section primarily contrasts our approach with other prominent works in the field. Initially, we consider video editing endeavors such as Pix2Pix Video, Text2LIVE, and Tune-a-Video, which output entire videos and have achieved certain successes in macroscopic editing effects, such as style transfer and object replacement. However, their performance in tasks requiring the preservation of identity features is subpar; they not only fail to ensure identity preservation but also alter most of the facial details.

\begin{figure}[htbp]

\begin{tabular}{@{}p{0.2\linewidth}@{}p{0.4\linewidth}@{}}

\begin{minipage}{0.2\textwidth} 
        Ours
\end{minipage}  & 
\begin{minipage}[c]{\textwidth} 
\includegraphics[width=0.4\linewidth]{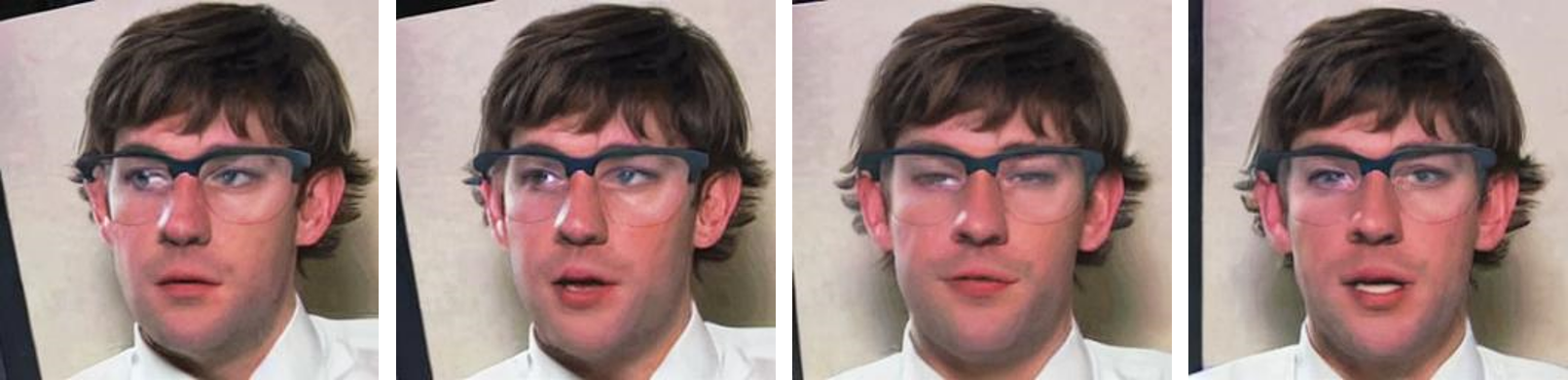} 
\end{minipage} \\
\addlinespace

\begin{minipage}{0.2\textwidth} 
Pix2Pix
\end{minipage}  & 
\begin{minipage}[c]{\textwidth} 
\includegraphics[width=0.4\linewidth]{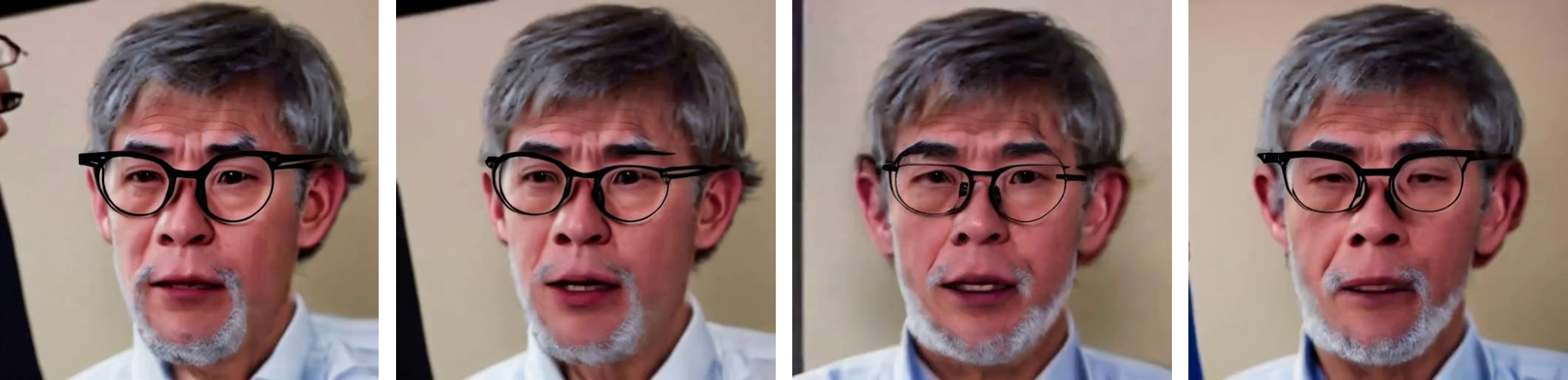} 
\end{minipage} \\
\addlinespace

\begin{minipage}{0.2\textwidth} 
LTX-Video
\end{minipage}  & 
\begin{minipage}[c]{\textwidth} 
\includegraphics[width=0.4\linewidth]{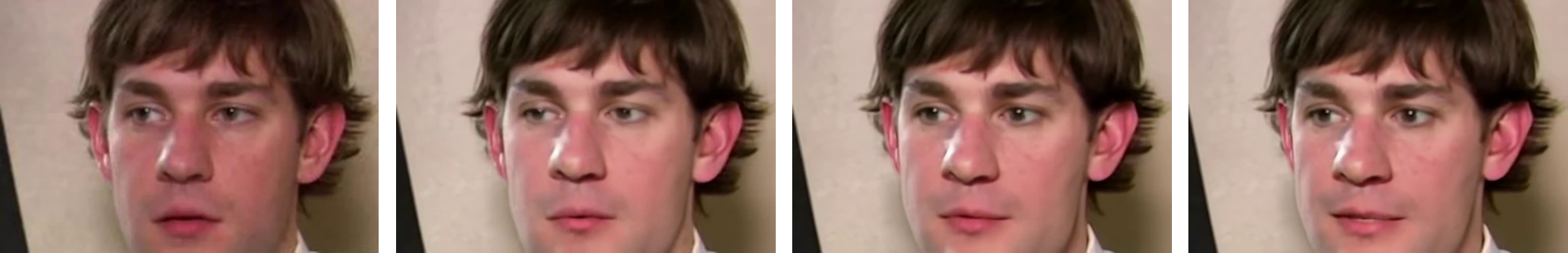} 
\end{minipage} \\
\addlinespace

\begin{minipage}{0.2\textwidth} 
SDXL SVD
\end{minipage}  & 
\begin{minipage}[c]{\textwidth} 
\includegraphics[width=0.4\linewidth]{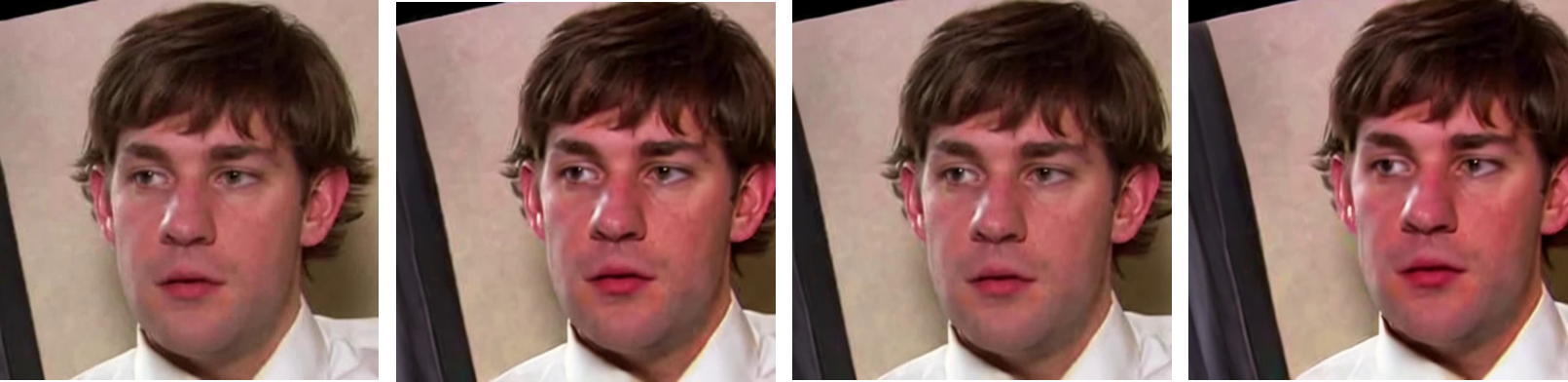} 
\end{minipage} \\
\addlinespace

\begin{minipage}{0.2\textwidth} 
AI VC
\end{minipage}  & 
\begin{minipage}[c]{\textwidth} 
\includegraphics[width=0.4\linewidth]{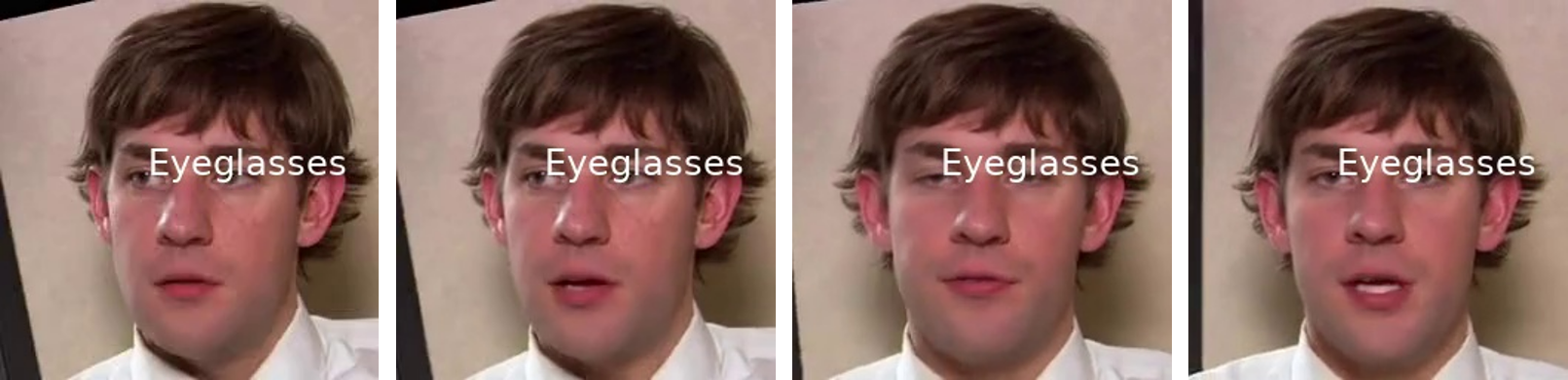}
\end{minipage} \\

\end{tabular}
\caption{Quality Comparison with other video editing or generation works.}

\label{fig:suppcmp}
\end{figure}

Subsequently, we examine works that generate videos based on text and images, such as SDXL SVD, which primarily generates video segments based on reference images, and LTX-Video, which incorporates additional prompt inputs. In our tested examples, these methods do not effectively incorporate editing semantics onto the faces and result in video content alterations that lose comparability. Some video creation works also fail to integrate visual semantics into the video content. Figure~\ref{fig:suppcmp} illustrates some relevant results.

\end{document}